\pdfoutput=1

\documentclass[11pt]{article}
\usepackage{coling2020}
\usepackage{times}
\usepackage{url}
\usepackage{latexsym}

\usepackage{textcomp}
\usepackage{fixltx2e}
\usepackage{graphicx}
\usepackage{amsmath}
\usepackage[ruled,vlined]{algorithm2e}
\graphicspath{ {figs/} }

\renewcommand{\textrightarrow}{$\rightarrow$}

\usepackage{booktabs}
\usepackage{multirow}
\usepackage{color}
\usepackage{titlesec}

\colingfinalcopy 

\title{Finding Prerequisite Relations between Concepts using Textbook}

\author{
Shivam Pal \\ Department of Electrical Engineering \\  Indian Institute of Technology, Kanpur \\ {\tt shivampa@iitk.ac.in}
        \And
Vipul Arora \\ Department of Electrical Engineering \\  Indian Institute of Technology, Kanpur \\ {\tt vipular@iitk.ac.in}
         \AND
Pawan Goyal \\ Department of Computer Science and Engineering \\  Indian Institute of Technology, Kharagpur \\ {\tt pawang@cse.iitkgp.ac.in} 
}

\date{}

\begin{document}
\maketitle
\begin{abstract}
 A \textit{prerequisite} is anything that you need to know or understand first before attempting to learn or understand something new. In the current work, we present a method of finding prerequisite relations between concepts using related textbooks. Previous researchers have focused on finding these relations using Wikipedia link structure through unsupervised and supervised learning approaches. In the current work, we have proposed two methods, one is statistical method and another is learning-based method. We mine the rich and structured knowledge available in the textbooks to find the content for those concepts and the order in which they are discussed. Using this information, proposed statistical method estimates explicit as well as implicit prerequisite relations between concepts. During experiments, we have found performance of proposed statistical method is better than the popular \textit{RefD method}, which uses Wikipedia link structure. And proposed learning-based method has shown a significant increase in the efficiency of supervised learning method when compared with \textit{graph} and \textit{text-based} learning-based approaches.
\end{abstract}

\section{Introduction}
\begin{figure}[htp]
    \centering
    \includegraphics[width = 7cm]{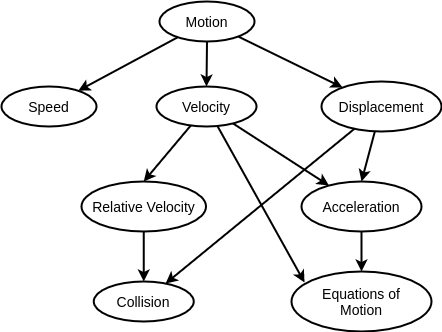}
    \caption{Prerequisite DAG, where `A' \textrightarrow `B' represents concept \textit{A} as a prerequisite of concept \textit{B}}
    \label{fig:prereq_dag}
\end{figure}

Nowadays, Intelligent Tutoring Systems (ITS) are gaining popularity, and there are a lot of researchers who are trying to build ITS for various applications \cite{Almasri2019-ALMITS}. Organising the \textit{Domain Knowledge} in these systems is crucial, and is achieved by arranging all the concepts to be taught in a \textit{Directed Acyclic Graph} (DAG), where concepts are present at nodes and an edge is the direction of prerequisite relation between pairs. For example as shown in Figure \ref{fig:prereq_dag}, \textit{velocity} and \textit{acceleration} are prerequisites of \textit{equations of motion}. It means, first we have to study \textit{velocity} and \textit{acceleration} before starting \textit{equations of motion}, i.e., the ITS system will recommend \textit{velocity} and \textit{acceleration} before teaching \textit{equations of motion}. ITS also uses prerequisite information in planning a personalised curriculum for the student and also modifies the curriculum based on their performance \cite{jeong2012personalized}. Prerequisite relations suggested by ITS are especially useful for making the self-guided online learning efficient, where readers are faced with a large amount of educational resources \cite{pan2017prerequisite} but have limited time.

The current work focuses on the problem of finding prerequisite relations between a given pair of concepts (\textit{A}, \textit{B}) from a textbook of the concerned subject. Mostly, previous works do not use textbooks, but only the Wikipedia data. Some researchers \cite{talukdar2012crowdsourced,liang2015measuring} have approached this problem using Wikipedia link structure by training the statistical measures which can predict the prerequisite relation. Other approaches \cite{wang2016using,liang2018investigating} create graph-based and text-based features from Wikipedia corpus and try to solve this problem through supervised learning. However, the lack of large scale prerequisite relation labels remains a major obstacle for effective machine-learning based solutions.

As far as we know, the current work is the first attempt to find prerequisite relations between concepts from textbooks. We use the rich information available in multiple textbooks in the form of table of contents and chapters' text along with Wikipedia content to derive a couple of features, which correlate well with the prerequisite relationships. 
We use simple threshold based method as well as supervised classification methods using the proposed features.

During experiments, it is found out that the proposed threshold-based method performs better than the popularly used RefD Method \cite{liang2015measuring}, which uses Wikipedia-link structure to predict the relations. Moreover, the proposed features improve the efficiency of supervised learning method when used with other \textit{graph-based} and \textit{text-based} features.

\section{Related Work}
Finding prerequisite relations is quite a new research area. Nevertheless, there is much data-driven research using different kinds of educational material and Wikipedia information.

Some approaches \cite{vuong2011method,scheines2014discovering,chen2016joint} try to find prerequisite relations using the students' performance data from different items. Such methods require an extensive amount of data which further needs to be processed. But such methods are not scalable, and annotating data is also a tedious task.

Some other approaches \cite{chaplot2016data,liu2016learning,pan2017prerequisite} try to find prerequisite relations using MOOC datasets. These methods process the online video content into text and use the internal links between the videos of courses.

Along with the MOOC dataset, there are works \cite{talukdar2012crowdsourced,liang2015measuring,wang2016using,liang2017recovering,liang2018investigating} that exploit Wikipedia dataset for finding prerequisite relations. Researchers have proposed both supervised learning methods, as well as unsupervised learning methods. In some cases, supervised learning methods outperform unsupervised learning methods but require a large amount of annotated data.

The work by \newcite{gordon2016modeling} tries to represent the scientific literature as a labelled graph, where nodes represent documents, concepts and metadata, and labelled edges represent relations between nodes using Latent Dirichlet Allocation.  \newcite{labutov2017semi} have proposed an approach which can find prerequisite relations from the textbook information using the probabilistic graphical model to construct a prerequisite classifier.

\section{Proposed Method}

Before we describe our proposed method, we briefly discuss the notations. 
Given $C=\{c_1,c_2,...,c_n\}$ as the set of $n$ concepts in a particular domain, our final task is to obtain a concept prerequisite matrix $\Omega=C^2 \rightarrow \{0,1\}$, where $(c_i, c_j) = 1$ indicates $c_j$ is prerequisite of $c_i$ and 0 indicates $c_j$ is not prerequisite of $c_i$.

For this task, we make use of Wikipedia content as well as textbook data. Let $W$ and $B$ denote the Wikipedia and textbook content. Thus, $W=\{w_1,w_2,...,w_n\}$, where $w_i$ is the Wikipedia content related to concept $c_i$. From the textbook, we make use of Table of content (ToC) section ($BS$),  ToC titles ($BT$) and book chapters ($BC$); thus $B=[BS,BT,BC]$. From this book, we specifically make use of

\begin{itemize}
    \item $\sigma=\{\sigma_1,\sigma_2,...,\sigma_n\}$ where $\sigma_i$ is the content related to $c_i$ in textbook $B$
    \item $\rho=\{\rho_1,\rho_2,...,\rho_n\}$ where $\rho_i$ is the positioning of $c_i$ in ToC section, $BT$
\end{itemize}

\subsection{Overview of the Proposed Method}
\begin{figure}[htp]
    \centering
    \includegraphics[width=14cm]{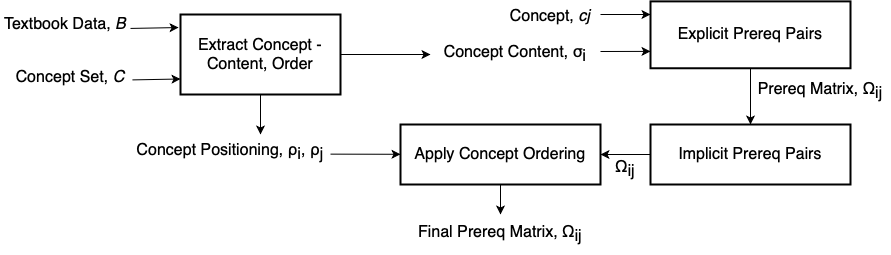}
    \caption{Proposed Method Overview}
    \label{fig:prop_method_overview}
\end{figure}

Figure \ref{fig:prop_method_overview} describes the overview of the proposed framework. Starting from the textbook data, $B$ and concept set, $C$, we first calculate $\sigma$ and $\rho$ in Section \ref{sec:content_order}.

In our current work we have proposed two features, \textit{book\_tfidf} and \textit{order\_diff} for making prerequisite relationship learning classification. For deriving these features, we have used the information stored in prerequisite matrix $\Omega$ and concept positions $\rho$. The method of finding these parameters has shown in figure \ref{fig:cont_ord_pipeline}.

We have started with textbook data, $B$ and concept set, $C$ and calculated $\sigma$ and $\rho$ in section \ref{sec:content_order}. In section \ref{sec:explicit_pairs} using $\sigma_j$, we  calculate the importance of concept $c_i$ in concept $c_j$ using TF-IDF measure and extract relations between explicit defined pairs. After that  in section \ref{sec:implicit_pairs}, we calculate implicit relations between concepts using the \textit{transitive property}, if concept $c_j$ is a prerequisite of concept $c_i$, and concept $c_k$ is a prerequisite of concept $c_j$, then we can say concept $c_k$ is a prerequisite of concept $c_i$. Finally we have removed wrong predicted pairs using concept ordering, $\rho$ in section \ref{sec:apply_concept_order}. In section \ref{sec:proposed_feature}, we have discussed our proposed features.

For example, there is a concept pair ($c_i$ = \textit{acceleration}, $c_j$ = \textit{velocity}) and we have to find whether $c_j$ is a prerequisite of $c_i$ or not. So we first extract content and position of  $c_i$ \& $c_j$ from textbook, then find the \textit{tfidf score} of $c_j$ in $\sigma_i$ (concept content of $c_i$). If $c_j$ is not present in $\sigma_j$ then we try to find it using \textit{transitive property}, and we get a value for pair ($c_i$, $c_j$). After that, we apply concept ordering on that pair as - if position of $c_j$ is after $c_i$ then make value of pair equals to zero, else keep it same. Finally by using the method discussed in section \ref{sec:proposed_method}, we classify whether \textit{velocity} is a prerequisite of \textit{acceleration} or not.

\subsubsection{Prior Knowledge}
\label{sec:prior_knowledge}




\paragraph{TFIDF Score}
\label{sec:tfidf}
We have used following method for finding tfidf where, \textit{c} is the concept from \textit{C}, $\sigma$ is the document, \textit{f} is the frequency of occurrence of \textit{c} in document $\sigma$, \textit{f\textsuperscript{'}} is the total number of concepts in $\sigma$, \textit{df} is the number of documents in which concept \textit{c} is occurring, and \textit{N} is the total number of concepts in \textit{C}.

\[
    \text{TF-IDF }(c,\text{ } \sigma) = \frac{f}{f\textsuperscript{'}}\times\log \frac{N}{df + 1}
\]

\paragraph{Document Matching using TFIDF Vectorizer}
\label{sec:tfidf_vectorizer}
For matching a document with another, we first convert each document into a vector by making collection of tokens as its dimension and its TFIDF score as value for that dimension then use \textit{cosine similarity} to measure the similarity between two documents by taking into account their \textit{vector space}



\subsubsection{Extract Concept - Content and Order}
\label{sec:content_order}
\begin{figure}[htp]
    \centering
    \includegraphics[width=14cm]{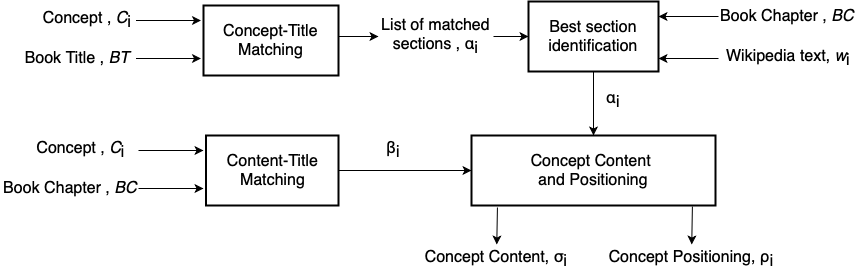}
    \caption{Concept - Content and Ordering Pipeline}
    \label{fig:cont_ord_pipeline}
\end{figure}

Our goal is to find $\sigma_i$, content of the concept and its position $\rho_i$ in the textbook \textit{B} related to concept $c_i$. We have followed the pipeline shown in figure \ref{fig:cont_ord_pipeline}.


We first match the concept $c_i$ with the textbook ToC titles \textit{BT} in section \ref{sec:match_title_concept} which contain content for concept $c_i$. Further in section \ref{sec:identify_best_sec}, we try to find best section where the concept is actually discussed in detail. We also need to extract sections by matching concepts with book chapter, discussed in section \ref{sec:concept_content_match}. Finally, using the data from section \ref{sec:identify_best_sec} and \ref{sec:concept_content_match}, we find $\sigma_i$ and $\rho_i$ in section \ref{sec:final_cc_co}.


\paragraph{Matching Concepts with Book ToC Title}
\label{sec:match_title_concept}
In this section, our goal is to match concept $c_i$ with $bt_j$ where i $\in$ $\{1, ..., n\}$ and j $\in$ $\{1, ..., m\}$. Let's introduce a variable $\alpha$ = \{$\alpha$\textsubscript{1}, $\alpha$\textsubscript{2}, ..., $\alpha$\textsubscript{n}\} which stores the list of sections $bs_j$ after matching $c_i$ with $bt_j$

\begin{center}
    $\alpha_i$ = \{$bs_j$ $|$ if ($c_i$ = $bt_j$ or $c_i$ in $bt_j$ or $bt_j$ in $c_i$), j $\in \{1, ..., m\}$\}    
\end{center}

\paragraph{Identifying Best Section for Concept}
\label{sec:identify_best_sec}
From the previous section, $\alpha$\textsubscript{i} may contain various sections which may have following two kinds of ambigutiy. In case, we don't get any section in $\alpha_i$, we treat that concept, $c_i$ as \textit{basic concept}, whose explanation is present in some other prerequisite book. Our goal is to remove following ambiguity if present in $\alpha_i$.

\begin{enumerate}
    \item \textbf{Hierarchical Ambiguity}: This kind of ambigutiy contains both parent and child sections. For example, in this list [3, 3.1, 3.2.2], both '3.1' and '3.2.2' are child of '3'. Here '3' will contain '3.1' and '3.2.2', so we have to choose most appropriate one
    \item \textbf{Multi-Chapter Ambiguity} This kind of ambiguity contains sections from different chapters. For example, in this [5.1, 7.2, 8.1], but we need only one section because concept is discussed only once in the textbook
\end{enumerate}

\textbf{Resolving Hierarchical Ambiguity}:
For resolving this ambiguity in $\alpha_i$, firstly we make hierarchical clusters (cluster of sections which lie in same chapter) in $\alpha_i$. Then for each cluster, we compare the book chapter \textit{BC} for each section in cluster with  Wikipedia content $w_i$ of concept $c_i$ using \textit{TFIDF Vectorizer Method}. Finally the section is returned from each cluster and $\alpha$\textsubscript{i} is updated.

\textbf{Resolving Multi-Chapter ambiguity}:
For resolving this ambiguity, we match book content of each section present in $\alpha_i$ with corresponding Wikipedia content $w_i$ using \textit{TFIDF Vectorizer Method} and return the section which has the highest similarity, then saved in $\alpha_i$. In this way we get a unique section for each concept $c_i$ stored in $\alpha_i$



\paragraph{Matching Concepts with Book Chapter}
\label{sec:concept_content_match}
In the textbook, concepts are discussed in an ordered manner such that the basic concepts are discussed first which further used in explaining higher order concepts. Let's define, $\beta$ = \{$\beta$\textsubscript{1}, $\beta_2$, ..., $\beta$\textsubscript{n}\}, where $\beta_i$ contains the section where it has been first discussed in textbook, $B$

\[
    \beta_{i} = bs_{j}, \text{ s.t. } j=\arg\min_{j'}\{ j'\in \{1,...,m\} : freq(c_{i}; bc_{j'}) > 1 \}
\]

Here $freq(a;b)$ is the frequency of string $a$ in text $b$. In experiments, we find if $freq(a;b)$ = 1 then concept is refer there in example or reference form, and if $freq(a;b)$ greater than 1 then that concept is discussed or used there in explaining higher concepts.

\paragraph{Final Concept Content and Positioning}
\label{sec:final_cc_co}
Finally, $\sigma_i$ and $\rho_i$ are calculated using the information stored in $\alpha_i$ and $\beta_i$. Here, match($A_i$, $B_i$) which matches content of section $A_i$ and $B_i$ with $w_i$ and return the best match using \textit{TFIDF Vectorizer} document matching method,

\[
    \rho_i=
    \begin{cases}
        \alpha_i,& \text{if }\beta_i \text{ is\_empty()}\\
        \text{match}(\beta\textsubscript{i}, \alpha_i),& \text{otherwise}
    \end{cases}
\]

\[
    \alpha_i=
    \begin{cases}
        \text{Empty},& \text{if }\alpha_i \text{ is\_empty()}\\
        \text{match}(\beta_i, \alpha_i),& \text{otherwise}
    \end{cases}
\]

\textit{Concept Content}($\sigma_i$): is the textbook content for section stored in $\alpha_i$


\subsubsection{Extracting Explicit Prerequisite Relations}
\label{sec:explicit_pairs}
Using the concept content $\sigma_i$, we have to calculate the confidence of having prerequisite relation between pair ($c_i$, $c_j$) (where $c_j$ is prerequisite of $c_i$) using \textit{TF-IDF Method} (discussed in \ref{sec:tfidf}) and update $\Omega$ as follows, where $\Omega_{ij}$ stores the value of ($c_i$, $c_j$)

\[
    \Omega\textsubscript{ij} = 
    \begin{cases}
        \text{tf-idf }(\textit{c}\textsubscript{j},\text{ } \sigma\textsubscript{i}),& \text{if }\textit{c}\textsubscript{j} \text{ is in } \sigma\textsubscript{i}\\
        0,& \text{otherwise}
    \end{cases}
\]


\subsubsection{Extracting Implicit Prerequisite Relations}
\label{sec:implicit_pairs}
There is also implicit relation exist between concepts in textbook,
for example, \textit{Linear Algebra} is a prerequisite of \textit{Neural Networks} but in the text \textit{Linear Algebra} is not explicitly mentioned but \textit{Matrix Multiplication} mentioned. For extracting these implicit relations, we use the hypothesis that, if concept $c_k$ is a prerequisite of concept $c_i$, and concept $c_j$ is a prerequisite of concept $c_k$; then we can say that concept $c_j$ is a prerequisite of concept $c_i$. Applying this heuristic in $\Omega$,

\begin{center}

$\Omega_{ij}$ = $\underset{k}{\operatorname{argmax}}$ (${\operatorname{argmin}}$ ($\Omega_{ik}$, $\Omega_{kj}$) ; $k \in \{1, ..., n\}$)

\end{center}

\subsubsection{Apply Concept Ordering}
\label{sec:apply_concept_order}
If the position $\rho_i$ (for concept $c_i$) is before the position $\rho_j$ (for concept $c_j$) then $c_j$ will never be prerequisite of $c_i$ but $c_i$ can be prerequisite of $c_j$. Apply this heuristic in $\Omega$ as follows 
\[
    \Omega_{ij} = 
    \begin{cases}
        \Omega_{ij},& \text{if }\rho_i > \rho_j \text{; } i,j \in \{1, ..., n\}\\
        0,& \text{otherwise}
    \end{cases}
\]

\subsubsection{Proposed Features}
\label{sec:proposed_feature}
Using the information, $\Omega$ and $\rho$, we propos two features, 1. \textit{book\_tfidf} ($c_i$, $c_j$), and 2. \textit{order\_diff} ($c_i$, $c_j$) for making a prerequisite learning classification model for given a pair of concepts ($c_i$, $c_j$) and we have to find whether $c_j$ is a prerequisite of $c_i$ or not

\begin{center}
    \textit{book\_tfidf}$(c_i, c_j)$ = $\Omega_{ij}$
\end{center}

\begin{center}
    \textit{order\_diff}$(c_i, c_j)$ = \textit{rank}$(c_i)$ - \textit{rank}$(c_j)$
\end{center}

where \textit{rank}$(c)$ is calculated by ordering all the concepts in $C$ based on their sections stored in $\rho$

\subsection{Proposed Method for Prerequisite Classification}
\label{sec:proposed_method}
We are trying to solve the problem of finding prerequisite relation between pair ($c_i$, $c_j$) and classify whether $c_j$ is a prerequisite of $c_i$ or not. Let's define the triplet, ($c_i, c_j, r_{ij}$) where $r_{ij} = 1$ if $c_j$ is a prerequisite of $c_i$, else $r_{ij} = 0$. For developing prerequisite classifier, we propose two methods, one is \textit{statistical method} (discussed in section \ref{sec:proposed_statisitcal_method}) and the other one is \textit{learning-based method} (discussed in section \ref{sec:proposed_learning_method}).

\subsubsection{Proposed Statistical Method}
\label{sec:proposed_statisitcal_method}
We use the prerequisite matrix $\Omega$ to find the prerequisite relations. Here we introduce the threshold parameter $\theta$, and based on the value $\theta$, we determine $r_{ij}$ as follows

\[
    r_{ij} = 
    \begin{cases}
        1,& \text{if } \Omega_{ij} > \theta\\
        0,& \text{otherwise}
    \end{cases}
\]

If the annotated datset is not available then in that case we can manually fine-tune the threshold parameter $\theta$ and if we have given the annotated dataset then we can calculate $\theta$ by plotting \textit{F1-Score} for various thresholds, $\theta \in \{0.0, 0.02, 0.04, ..., 1.0\}$ over training dataset pairs and selected the $\theta$ for which we which we get highest \textit{F1-Score}.

The proposed statistical method does not need labelled data for classification and therefore we don't face bootstrap problem for a new course or a new subject.


\subsubsection{Proposed Learning-Based Method}
\label{sec:proposed_learning_method}
In this learning based method, we employ supervised learning approach, where we have given a pair of concept ($c_i$, $c_j$) and we have to train a classifier which can output $r_{ij} = \{0, 1\}$, where $1$ represents $c_j$ is prerequisite of $c_i$ and $0$ represents $c_j$ is not prerequisite of $c_i$. For the concept pair ($c_i$, $c_j$), we use \textit{graph-based} and \textit{text-based} features \cite{liang2018investigating} for training the model. We also append the input with the \textit{proposed} features.
\begin{itemize}
    \item \textbf{Graph-based} features: In/Out Degree, Common Neighbors, \#Links, Link Proportion, Normalised Google Distance, Pointwise Mutual Information, Reference Distance, Page-Rank, Hyperlink Induced Topic Search
    
    \item \textbf{Text-based} features: 1st Sentence, In Title, Title Jaccard Similarity, Length, Mention, Noun Phrases, TF-IDF Similarity, Word2vec Similarity, LDA Entropy, LDA Cross Entropy
    
    \item \textbf{Proposed} features: book\_tfidf, order\_diff
\end{itemize}

For the binary classification problem at hand, we employ Random Forest, Support Vector Machines, Logistic Regression and Naive Bayes methods.

\section{Experimental Settings}

\subsection{Dataset Collection}

For our experiments, we utilize the data from three domains - \textit{Geometry}\footnote{Geometry: Dan Greenberg, Lori Jordan, Andrew Gloag, Victor Ci- farelli, Jim Sconyers,Bill Zahnerm, ”CK-12 Basic Geometry”}, \textit{Physics}\footnote{Mark Horner, Samuel Halliday, Sarah Blyth, Rory Adams, Spencer Wheaton, ”Textbooks for High School Students Studying the Sciences”, 2008} and \textit{Precalculus}\footnote{Stewart, James, Lothar Redlin, and Saleem Watson. "Precalculus: Mathematics for calculus". Cengage Learning, 2015}. To construct the final dataset, we require following data for each domain - 
1) \textit{Labeled Pairs ($c_i$, $c_j$, $r_{ij}$)}: where $r_{ij}$ = 1, if concept $c_j$ is a prerequisite of concept $c_i$ else $r_{ij}$ = 0, this dataset is avaialble online\footnote{https://github.com/harrylclc/AL-CPL-dataset}; 
2) \textit{Wikipedia Content for Concepts} contains content corresponding to each concept $c_i$, this we have extracted from April, 2020 Wikipdeia dump; 
3) \textit{Concept Synonyms} contains the synonym terms for concepts in textbook, this we have collected from \cite{wang2015concept} work and available online\footnote{https://github.com/dayouzi/CHEB}; and cleaned it manually wherever it is required; and last 
4) \textit{Book Dataset} contains the complete data (ToC titles, ToC sections and content in each section) of textbook, this is collected manually from the specified books.

Table \ref{tab:data_stats} contains the statistics of the labelled pairs and concept present in each domain. Before using the book dataset, we normalise the concept terms using concept synonyms data.

\begin{table*}[htp]
\centering
\begin{tabular}{lllll}
\hline
\textbf{Domain} & \textbf{\#Concepts} & \textbf{\#Pairs} & \textbf{\#Positive Pairs} & \textbf{\#Negative Pairs}\\
\hline
Geometry & 89 & 1681 & 524 & 1154\\
Physics & 152 & 1962 & 487 & 1475\\
Precalculus & 113 & 918 & 338 & 580\\
\hline
\end{tabular}
\caption{Prerequisite Dataset Statistics}
\label{tab:data_stats}
\end{table*}

\subsection{Baseline}
We have used the popular statistical method, RefD and supervised learning-based method for comparing the efficiency of proposed method.

\begin{enumerate}
    \item \textbf{Reference Distance}: We employ Reference Distance (RefD) as one of our baselines. It is a Wikipedia-link based method which finds the prerequisite relation between pairs of concept by measuring how two different concepts refer to each other. This method is only applicable to Wikipedia concepts. To make this method comparable, we have calculated RefD between each labelled pair of concepts using both \textit{Equal Method} as well as \textit{TFIDF Method}.
    
    \item \textbf{Supervised Learning}: We employ the method of supervised learning as described in the paper \cite{liang2018investigating} as our another baseline. This method uses \textit{Graph-based} and \textit{Text-based} features from Wikipedia Content and then train the supervised model with these features. This trained model predicts the prerequisite relation between concepts ($c_i$, $c_j$). For our convenience, lets name this method as \textit{Graph and Text based features classification} learning method (\textit{GTC learning method})

\end{enumerate}


\section{Experimental Results}
For experiment purposes, we have used datasets from three domains - \textit{Geometry}, \textit{Physics} and \textit{Precalculus}. We have compared the efficiency of proposed statistical method with RefD method and proposed learning method with GTC learning method. We employ \textit{k-Fold Cross Validation} method for training and testing the model efficiency and take the average score from each fold. We have used \textbf{\textit{K = 5}}, i.e. \textit{5-Fold Cross Validation} for model training. We have used \textit{Precision} (\textit{P}), \textit{Recall} (\textit{R)}, \textit{F1-Score} (\textit{F1}) and \textit{Area under Precision-Recall Curve} (\textit{AUPRC}) for model comparison.

\begin{figure}[htp]
    \centering
    \includegraphics[width = 15cm]{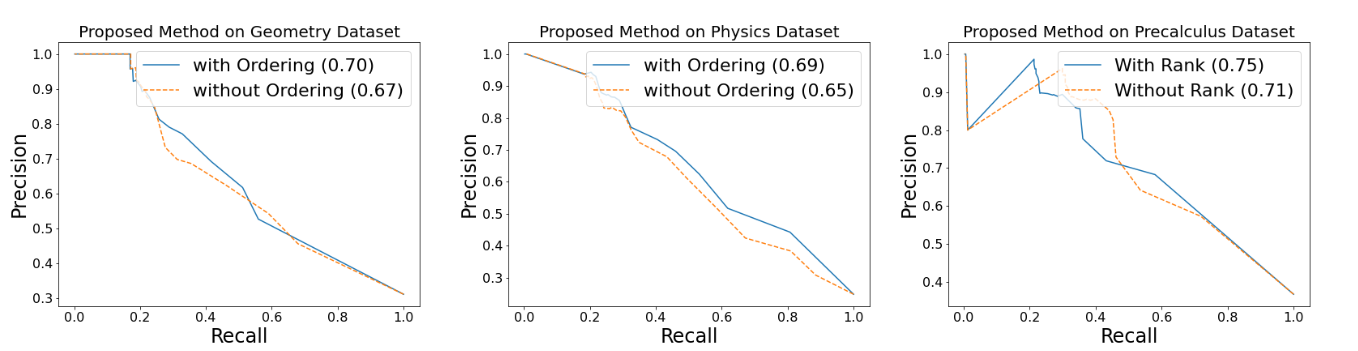}
    \caption{Area under Precision-Recall Curve for Proposed Statistical Method}
    \label{fig:proposed_method}
\end{figure}

\subsection{Proposed Statistical Method}
In figure \ref{fig:proposed_method}, we have shown AUPRC for proposed statistical method with a \textit{solid} line and along with that, we experimented the impact of concept ordering (as discussed in section \ref{sec:apply_concept_order}). Results without applying concept ordering has been shown with \textit{dashed} line. As we can clearly see in figure \ref{fig:proposed_method} that efficiency of proposed statistical method has significantly increased after applying concept ordering.
For the proposed statistical method, we get threshold values as $\theta$ = 0.06 for geometry, $\theta$ = 0.12 for physics and $\theta$ = 0.04 for precalculus.

\subsection{Statistical Method Comparison}
\begin{figure}[htp]
    \centering
    \includegraphics[width = 16cm]{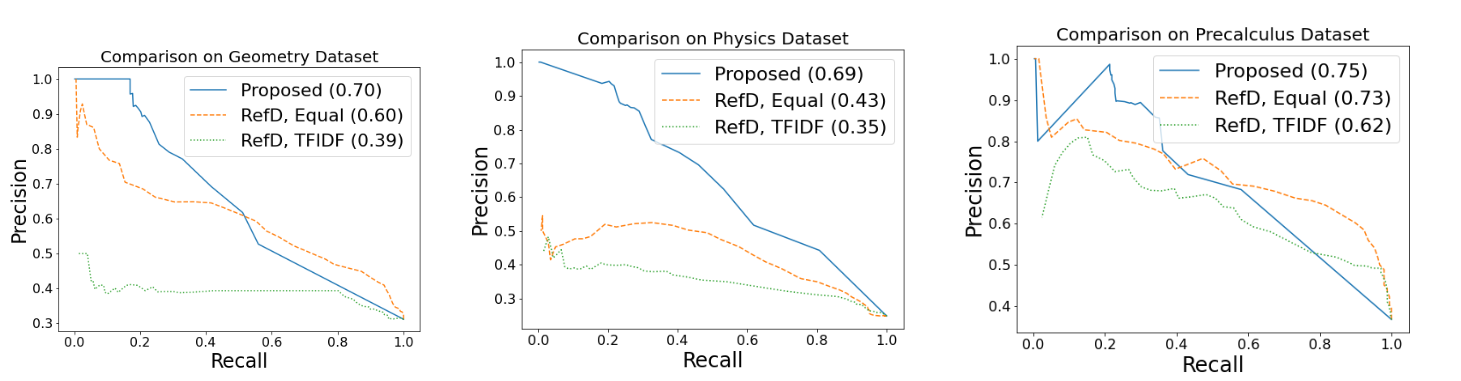}
    \caption{Area under Precision-Recall Curve comparison between RefD and Proposed Statistical Method}
    \label{fig:unsup_comparison}
\end{figure}

In figure \ref{fig:unsup_comparison}, we have compared the efficiency of proposed statistical method with the popular RefD method ($w$ = \textit{equal} and \textit{tfidf}) using AUPRC. From the figure \ref{fig:unsup_comparison}, we can see that AUPRC is higher for proposed statistical method in all the datasets than RefD method. In the table \ref{tab:unsup_comapre}, we have compared propsed method with RefD method over various parameters. We have found - 1) proposed method is giving higher \textit{AUPRC} than RefD Method, 2) proposed method is giving higher \textit{Precision} and lower \textit{Recall}, 3) RefD method is giving lower \textit{Precision} and higher \textit{Recall}. We get higher \textit{F1-Score} with proposed method for geometry and physics datasets, but lesser for precalculus. The reason may be that we have less amount of labelled pairs in case of precalculus for comparison.

\begin{table*}[htp]
    \centering
    \begin{tabular}{llllllllllll}
  \toprule
  Domain & Measure & Proposed & RefD (Equal) & RefD (TFIDF)\\
  \midrule
  
  \multirow{3}{*}{Geometry} & P & \textbf{62.2} & 50.6 & 39.3\\
  & R & 49.3 & 71.5 & \textbf{80.1}\\
  & F1 & 54.9 & \textbf{59.1} & 52.7\\
  & AUPRC & \textbf{0.70} & 0.60 & 0.38\\
  \midrule
  \multirow{3}{*}{Physics} & P & \textbf{60.8} & 42.1 & 32.2\\
  & R & 58.5 & 60.8 & \textbf{71.1}\\
  & F1 & \textbf{59.6} & 49.7 & 44.2\\
  & AUPRC & \textbf{0.69} & 0.43 & 0.35\\
  \midrule
  \multirow{3}{*}{Precalculus} & P & \textbf{68.6} & 62.1 & 54.0\\
  & R & 54.4 & 82.4 & \textbf{73.4}\\
  & F1 & 60.4 & \textbf{70.7} & 61.8\\
  & AUPRC & \textbf{0.75} & 0.73 & 0.61\\
  \bottomrule
\end{tabular}
\caption{Statistical Methods Efficiency Comparison}
\label{tab:unsup_comapre}
\end{table*}

\subsection{Learning-Based Method Comparison}
We employ four widely used binary classifiers - Random Forest (RF), Support Vector Machine (SVM), Logistic Regression (LR) and Naive Bayes (NB). For experiments, we set parameter as, C = 1.0 for LR, use a linear kernel for SVM, and use 200 trees for RF. We use \textit{5-Fold Cross Validation} for evaluating each dataset and report the average score.

We first experiment with \textit{GTC learning method} and report all the results in table \ref{tab:sup_compare}. In this experiment, Random Forest (RF) performed better than other models with highest F1-score for all the datasets. After that, we have performed the same experiment with \textit{proposed learning method} (discussed in section \ref{sec:proposed_learning_method}) and report all the results in table \ref{tab:sup_compare}. With proposed learning method also random forest performed better than other models.

After comparing the results of of both methods, we can see that precision, recall and f1-score are higher for proposed learning method and shows the overall efficiency in supervised learning-based approaches.

\begin{table*}[htp]
    \centering
    \begin{tabular}{llllllllllll}
  \toprule
  Domain & Measure &\multicolumn{4}{c}{GTC Learning Method} & & \multicolumn{4}{c}{Proposed Learning Method} \\
  \midrule
  && RF & SVM & LR & NB & & RF & SVM & LR & NB \\
  \cmidrule{3-6} \cmidrule{8-11}
  \multirow{3}{*}{Geometry} & P & \textbf{94.5} & 82.3 & 84.2 & 84.6 & & 94.4 & 83.6 & 84.8 & 84.8\\
  & R & 85.8 & 66.3 & 62.0 & 44.7 & & \textbf{88.6} & 69.0 & 64.7 & 44.5\\
  & F1 & 89.9 & 73.4 & 71.4 & 58.4 & & \textbf{91.4} & 75.5 & 73.3 & 58.3\\
  \midrule
  \multirow{3}{*}{Physics} & P & 82.6 & 77.4 & 78.2 & 54.0 & & \textbf{85.4} & 77.5 & 76.8 & 59.7\\
  & R & 62.1 & 52.1 & 48.3 & 72.4 & & \textbf{66.1} & 55.5 & 52.5 & 72.3\\
  & F1 & 70.8 & 62.2 & 59.6 & 61.6 & & \textbf{74.4} & 64.6 & 62.2 & 65.2\\
  \midrule
  \multirow{3}{*}{Precalculus} & P & 89.8 & 88.6 & 86.2 & 81.1 & & \textbf{90.9} & 89.0 & 85.9 & 81.1\\
  & R & 90.1 & 86.1 & 81.9 & 78.1 & & \textbf{90.3} & 87.5 & 83.2 & 76.3\\
  & F1 & 89.9 & 87.2 & 83.9 & 79.2 & & \textbf{90.5} & 88.2 & 84.4 & 78.3\\
  \bottomrule
\end{tabular}
\caption{Learning-Based Models Efficiency Comparison}
\label{tab:sup_compare}
\end{table*}


\section{Conclusion}
The problem we are trying to address is a relatively new research area and highly useful in adaptive and personalised learning environments. During experiments, it is found that the proposed statistical method performs better than the RefD method, which uses Wikipedia-link structure to predict the relations. Apart from this, our proposed learning method gives better efficiency than GTC learning method which is based on \textit{graph-based} and \textit{text-based} features.

The proposed method doesn't require manually annotated data which was the major drawback of supervised learning approaches. Our method gives features which can be used in unsupervised way or can be incorporated in supervised learning, if manual annotations are available. Also, there are many niche topics whose content is not available on Wikipedia, but available in textbooks. Moreover, our proposed method can work for languages other than English as well, that may not have rich Wikipedia data available.

\section{Future Work}
In the proposed work, we are finding the order of concepts from textbooks using the rule-based method, but it may be improved with the help of converting concepts into concept-vector space using the relative context of various concepts. Currently, the major drawback of supervised learning is that it doesn't perform well over cross-domains. We can think of \textit{self-learning} methods or \textit{transfer-learning} approaches for improving the performance of supervised learning over cross-domains.

We can extend our research in creating personalised curriculum planner system which asks students to enter the concepts they currently know and what they want to learn. Based on this knowledge, the system will create a personalised curriculum for them using their input information and prerequisite relations.

\bibliographystyle{coling2020}
\bibliography{coling2020}

\end{document}